\documentclass[11pt,a4paper]{article}
\usepackage[hyperref]{acl2021}
\usepackage{times}
\usepackage{latexsym}
\usepackage{amsfonts}
\usepackage{graphicx}
\usepackage{caption}
\usepackage{subfig}
\usepackage{amsmath}
\usepackage{multirow}
\usepackage{booktabs}
\usepackage{bm}
\usepackage{stfloats}

\usepackage{microtype}

\aclfinalcopy 
\def\aclpaperid{404} 

\title{Psycholinguistic Tripartite Graph Network for Personality Detection}

\author{Tao Yang, Feifan Yang, Haolan Ouyang, Xiaojun Quan\thanks{\; Corresponding author.}\\
  School of Computer Science and Engineering, Sun Yat-sen University, China \\
  \texttt{\{yangt225,yangff6,ouyhlan\}@mail2.sysu.edu.cn} \\
	\texttt{quanxj3@mail.sysu.edu.cn}}

\date{}

\begin{document}
\maketitle
\begin{abstract}
	Most of the recent work on personality detection from online posts adopts multifarious deep neural networks to represent the posts and builds predictive models in a data-driven manner, without the exploitation of psycholinguistic knowledge that may unveil the connections between one's language usage and his psychological traits.~In this paper, we propose a psycholinguistic knowledge-based tripartite graph network, \textit{TrigNet}, which consists of a tripartite graph network and a BERT-based graph initializer. The graph network injects structural psycholinguistic knowledge from LIWC, a computerized instrument for psycholinguistic analysis, by constructing a heterogeneous tripartite graph. The graph initializer is employed to provide initial embeddings for the graph nodes. To reduce the computational cost in graph learning, we further propose a novel flow graph attention network (GAT) that only transmits messages between neighboring parties in the tripartite graph.~Benefiting from the tripartite graph, TrigNet can aggregate post information from a psychological perspective, which is a novel way of exploiting domain knowledge.~Extensive experiments on two datasets show that TrigNet outperforms the existing state-of-art model by 3.47 and 2.10 points in average F1. Moreover, the flow GAT reduces the FLOPS and Memory measures by 38\% and 32\%, respectively, in comparison to the original GAT in our setting.
\end{abstract}

\section{Introduction}

Personality detection from online posts aims to identify one's personality traits from the online texts he creates. This emerging task has attracted great interest from researchers in computational psycholinguistics and natural language processing due to the extensive application scenarios such as personalized recommendation systems \cite{yang2019mining, jeong2020adaptive}, job screening \cite{hiemstra2019applicant} and psychological studies \cite{goreis2019systematic}. 

\begin{figure}[t]
	\centering
	\includegraphics[width=0.46\textwidth]{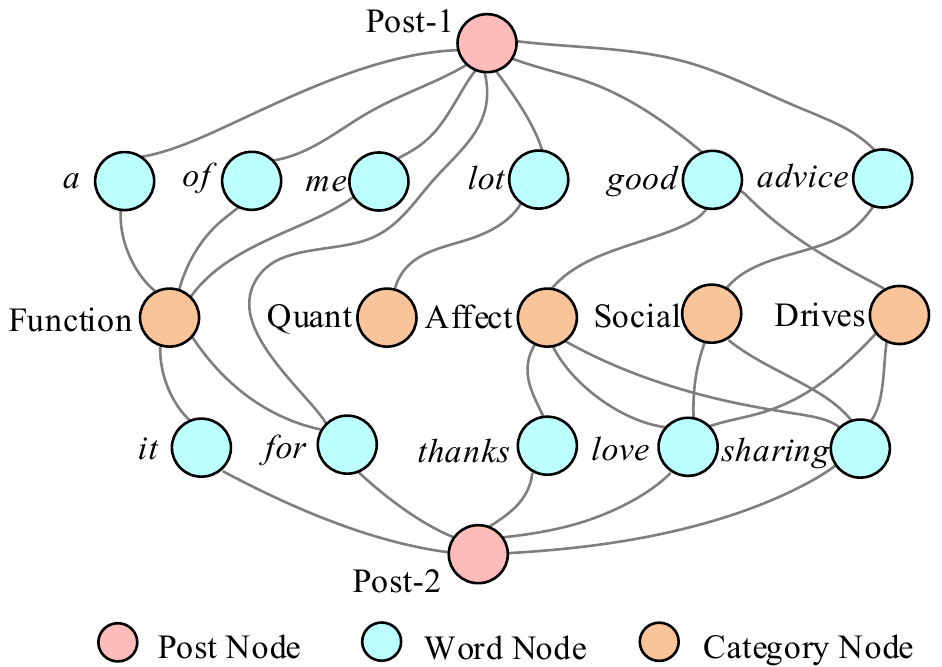}
	\caption{An example of our tripartite graph. The content of \emph{Post-1} and \emph{Post-2} are ``\textit{A lot of good advise for me.}" and ``\textit{Love it! Thanks for sharing!}", respectively.}
	\label{figure1}
	\vspace{-0.4cm}
\end{figure}

Psychological research shows that the words people use in daily life reflect their cognition, emotion, and personality \cite{gottschalk1997unobtrusive, golbeck2016predicting}. As a major psycholinguistic instrument, Linguistic Inquiry and Word Count (LIWC) \cite{tausczik2010psychological} divides words into psychologically relevant categories (e.g., \emph{Function}, \emph{Affect}, and \emph{Social} as shown in Figure \ref{figure1}) and is commonly used to extract psycholinguistic features in conventional methods \cite{golbeck2011predicting, sumner2012predicting}. Nevertheless, most recent works \cite{hernandez2017predicting,jiang2019automatic,keh2019myers,lynn2020hierarchical,gjurkovic2020pandora} tend to adopt deep neural networks (DNNs) to represent the posts and build predictive models in a data-driven manner.~They first encode each post separately and then aggregate the post representations into a user representation. Although numerous improvements have been made over the traditional methods, they are likely to suffer from limitations as follows.~First, the input of this task is usually a set of \emph{topic-agnostic} posts, some of which may contain few personality cues. Hence, directly aggregating these posts based on their contextual representations may inevitably introduce noise. Second, personality detection is a typical data-hungry task since it is non-trivial to obtain personality tags, while DNNs implicitly extract personality cues from the texts and call for tremendous training data. Naturally, it is desirable to explicitly introduce psycholinguistic knowledge into the models to capture critical personality cues. 

Motivated by the above discussions, we propose a psycholinguistic knowledge-based tripartite graph network, namely \textit{TrigNet}, which consists of a tripartite graph network to model the psycholinguistic knowledge and a graph initializer using a pre-trained language model such as BERT \cite{devlin2018bert} to generate the initial representations for all the nodes. As illustrated in Figure \ref{figure1}, a specific tripartite graph is constructed for each user, where three heterogeneous types of nodes, namely \emph{post}, \emph{word}, and \emph{category}, are used to represent the posts of a user, the words contained both in his posts and the LIWC dictionary, and the psychologically relevant categories of the words, respectively. The edges are determined by the subordination between \emph{word} and \emph{post} nodes as well as between \emph{word} and \emph{category} nodes. Besides, considering that there are no direct edges between homogeneous nodes (e.g., between post nodes) in the tripartite graph, a novel flow GAT is proposed to only transmit messages between neighboring parties to reduce the computational cost and to allow for more effective interaction between nodes. Finally, we regard the averaged post node representation as the final user representation for personality classification. Benefiting from the tripartite graph structure, the interaction between posts is based on \textit{psychologically relevant} words and categories rather than topic-agnostic context. 

We conduct extensive experiments on the Kaggle and Pandora datasets to evaluate our TrigNet model. Experimental results show that it achieves consistent improvements over several strong baselines. Comparing to the state-of-the-art model, SN+Att \cite{lynn2020hierarchical}, TrigNet brings a remarkable boost of 3.47 in averaged Macro-F1 (\%) on Kaggle and a boost of 2.10 on Pandora. Besides, thorough ablation studies and analyses are conducted and demonstrate that the tripartite graph and the flow GAT play an irreplaceable role in the boosts of performance and decreases of computational cost.

Our contributions are summarized as follows:
\begin{itemize}
	\item This is the first effort to use a tripartite graph to explicitly introduce psycholinguistic knowledge for personality detection, providing a new perspective of using domain knowledge.
	
	\item We propose a novel tripartite graph network, TrigNet, with a flow GAT to reduce the computational cost in graph learning.
	
	\item We demonstrate the outperformance of our TrigNet over baselines as well as the effectiveness of the tripartite graph and the flow GAT by extensive studies and analyses.
\end{itemize}


\section{Related Work}
\subsection{Personality Detection}
As an emerging research problem, text-based personality detection has attracted the attention of both NLP and psychological researchers \cite{cui2017survey,xue2018deep,keh2019myers,jiang2019automatic,tadesse2018personality,lynn2020hierarchical}.  

Traditional studies on this problem generally resort to feature-engineering methods, which first extracts various psychological categories via LIWC \cite{tausczik2010psychological} or statistical features by the bag-of-words model \cite{zhang2010understanding}. These features are then fed into a classifier such as SVM \cite{cui2017survey} and XGBoost \cite{tadesse2018personality} to predict the personality traits. Despite interpretable  features that can be expected, feature engineering has such limitations as it relies heavily on manually designed features.

With the advances of deep neural networks (DNNs), great success has been achieved in personality detection. \citet{tandera2017personality} apply LSTM \cite{hochreiter1997long} on each post to predict the personality traits. \citet{xue2018deep} develop a hierarchical DNN, which depends on an AttRCNN and a variant of Inception \cite{szegedy2016inception} to learn deep semantic features from the posts. \citet{lynn2020hierarchical} first encode each post by a GRU \cite{cho2014learning} with attention and then pass the post representations to another GRU to produce the whole contextual representations. Recently, pre-trained language models have been applied to this task.~\citet{jiang2019automatic} simply concatenate all the utterances from a single user into a document and encode it with BERT \cite{devlin2018bert} and RoBERTa \cite{liu2019roberta}. \citet{gjurkovic2020pandora} first encode each post by BERT and then use CNN \cite{lecun1998gradient} to aggregate the post representations. Most of them focus on how to obtain more effective contextual representations, with only several exceptions that try to introduce psycholinguistic features into DNNs, such as \citet{majumder2017deep} and \citet{xue2018deep}. However, these approaches simply concatenate psycholinguistic features with contextual representations, ignoring the gap between the two spaces. 

\subsection{Graph Neural Networks}
Graph neural networks (GNNs) can effectively deal with tasks with rich relational structures and learn a feature representation for each node in the graph according to the structural information. Recently, GNNs have attracted wide attention in NLP \cite{cao2019bag,yao2019graph,wang2020relational,wang2020heterogeneous}. Among these research, graph construction lies at the heart as it directly impacts the final performance. \citet{cao2019bag} build a graph for question answering, where the nodes are entities, and the edges are determined by whether two nodes are in the same document. \citet{yao2019graph} construct a heterogeneous graph for text classification, where the nodes are documents and words, and the edges depend on word co-occurrences and document-word relations. \citet{wang2020relational} define a dependency-based graph by utilizing dependency parsing, in which the nodes are words, and the edges rely on the relations in the dependency parsing tree. \citet{wang2020heterogeneous} present a heterogeneous graph for extractive document summarization, where the nodes are words and sentences, and the edges depend on sentence-word relations. Inspired by the above successes, we construct a tripartite graph, which exploits psycholinguistic knowledge instead of simple document-word or sentence-word relations and is expected to contribute towards psychologically relevant node representations.

\begin{figure*}[t]
	\centering
	\includegraphics[width=1\textwidth]{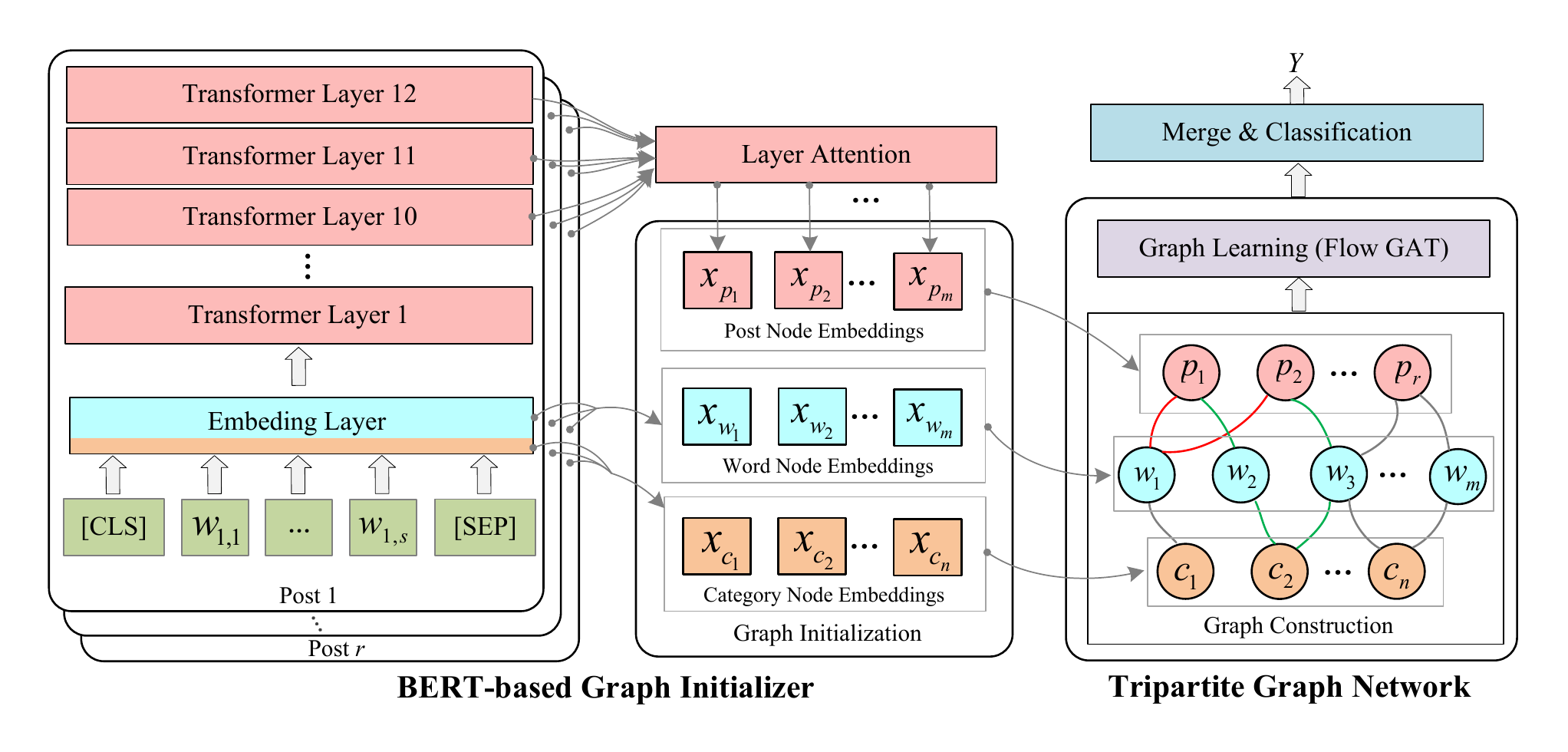}
	\caption{Overall architecture of our TrigNet, which consists of two modules: (1) a tripartite graph network (right) to inject psycholinguistic knowledge and (2) a BERT-based graph initializer (left) to initialize node embeddings.}
	\label{fig:architecture}
	\vspace{-0.3cm}
\end{figure*}

\section{Our Approach}
Personality detection can be formulated as a multi-document multi-label classification task \cite{lynn2020hierarchical,gjurkovic2020pandora}. Formally, each user has a set $P{\rm{=}}\left\{ {{p_1}, {p_2}, \ldots, {p_r}} \right\}$ of posts. Let ${p_i}{\rm{ = }}\left[ {{w_{i,1}}, {w_{i,2}}, \ldots, {w_{i,s}}} \right]$ be the $i$-th post with $s$ words, where $p_i$ can be viewed as a document.~The goal of this task is to predict $
T$ personality traits $Y{\rm{ = }}\left\{ {{y^t}} \right\}_{t = 1}^T$ for this user based on $P$, where $y^t \in \left\{ {0,1} \right\}$ is a binary variable.

Figure \ref{fig:architecture} presents the overall architecture of the proposed TrigNet, which consists of a tripartite graph network and a BERT-based graph initializer. The former module aims to explicitly infuse psycholinguistic knowledge to uncover personality cues contained in the posts and the latter to encode each post and provide initial embeddings for the tripartite graph nodes. In the following subsections, we detail how the two modules work in four steps: graph construction, graph initialization, graph learning, and merge \& classification.

\subsection{Graph Construction}
As a major psycholinguistic analysis instrument, LIWC \cite{tausczik2010psychological} divides words into psychologically relevant categories and is adopted in this paper to construct a heterogeneous tripartite graph for each user. 

As shown in the right part of Figure \ref{fig:architecture}, the constructed tripartite graph $\mathcal G {\rm{ = }} \left( {\mathcal V,\mathcal E} \right)$ contains three heterogeneous types of nodes, namely \emph{post}, \emph{word}, and \emph{category}, where $\mathcal V$ denotes the set of nodes and $\mathcal E$ represents the edges between nodes.~Specifically, we define $\mathcal V {\rm{ = }} {\mathcal V_p} \cup {\mathcal V_w} \cup {\mathcal V_c}$, where ${\mathcal V_p}{\rm{ = }}P{\rm{ = }}\left\{ {{p_1},{p_2}, \cdots ,{p_r}} \right\}$ denotes $r$ posts, ${\mathcal V_w}{\rm{ = }}\left\{ {{w_1},{w_2}, \cdots ,{w_m}} \right\}$ denotes $m$ unique psycholinguistic words that appear both in the posts $P$ and the LIWC dictionary, and ${\mathcal V_c}{\rm{ = }}\left\{ {{c_1},{c_2}, \cdots ,{c_n}} \right\}$ represents $n$ psychologically relevant categories selected from LIWC. The undirected edge $e_{ij}$ between nodes $i$ and $j$ indicates word $i$ either belongs to a post $j$ or a category $j$. 

The interaction between posts in the tripartite graph is implemented by two flows:~(1) ``$p {\rm{\leftrightarrow}} w {\rm{\leftrightarrow}} p$", which means posts interact via their shared psycholinguistic words (e.g., ``$p_1{\rm{\leftrightarrow}} w_1{\rm{\leftrightarrow}} p_2$" as shown by the red lines in Figure \ref{fig:architecture}); (2) ``$p {\rm{\leftrightarrow}} w {\rm{\leftrightarrow}} c {\rm{\leftrightarrow}} w {\rm{\leftrightarrow}} p$", which suggests that posts interact by words that share the same category (e.g., ``$p_1{\rm{\leftrightarrow}}w_2{\rm{\leftrightarrow}}c_2{\rm{\leftrightarrow}}w_3{\rm{\leftrightarrow}}p_2$" as shown by the green lines in Figure \ref{fig:architecture}). Hence, the interaction between posts is based on psychologically relevant words or categories rather than topic-agnostic context.

\subsection{Graph Initialization}
As shown in the left part of Figure \ref{fig:architecture}, we employ BERT \cite{devlin2018bert} to obtain the initial embeddings of all the nodes. BERT is built upon the multi-layer Transformer encoder \cite{vaswani2017attention}, which consists of a word embedding layer and 12 Transformer layers.\footnote{``BERT-BASE-UNCASED" is used in this study.}

\noindent \textbf{Post Node Embedding} \  The representations at the 12-th layer of BERT are usually used to represent an input sequence. This may not be appropriate for our task as personality is only weakly related to the higher order semantic features of posts, making it risky to rely solely on the final layer representations. In our experiments (Section \ref{sec:layer_attention}), we find that the representations at the 11-th and 10-th layers are also useful for this task. Therefore, we utilize the representations at the last three layers to 
initialize the post node embeddings. Formally, the representations ${x_{p_i}^j}$ of the $i$-th post at the $j$-th layer can be obtained by:
\begin{equation}
	{x_{p_i}^j} {\rm{=}} {{\rm{BERT}}^j}\left( {\left[ {{\rm{CLS}},{w_{i,1}}, \cdots ,{w_{i,m}},{\rm{SEP}}} \right]} \right) 
\end{equation}
where ``CLS'' and ``SEP'' are special tokens to denote the start and end of an input sentence, respectively, and
${\mathop{\rm BERT}^j\nolimits} \left(\cdot  \right)$ denotes the representation of the special token ``CLS" at the $j$-th layer. In this way, we obtain the representations ${\left[ {x_{{p_i}}^{10},x_{{p_i}}^{11},x_{{p_i}}^{12}} \right]^{\rm{T}}} \in {\mathbb{R}^{3 \times d}}$ of the last three layers, where $d$ is the dimension of each representation. We then apply layer attention \cite{peters2018deep} to collapse the three representations into a single vector $x_{{p_i}}$:
\vspace{-0.5cm}
\begin{equation} \label{eq2}
	{x_{{p_i}}} = \sum\limits_{j = 10}^{12} {{\alpha _j}x_{{p_i}}^j} 
\end{equation}
where $\alpha_j$ are softmax-normalized layer-specific weights to be learned. Consequently, we can obtain a set of post representations for the given $r$ posts of a user ${\bf X_p} = {\left[ {{x_{p_1}},{x_{p_2}}, \cdots ,{x_{p_r}}} \right]^{\rm{T}}} \in {\mathbb{R}^{r \times d}}$

\noindent \textbf{Word Node Embedding} \ BERT applies WordPiece \cite{wu2016google} to split words, which also cuts out-of-vocabulary words into small pieces. Thus, we obtain the initial node embedding of each word in $\mathcal V_w$ by considering two cases: (1) If the word is not out of vocabulary, we directly look up the BERT embedding layer to obtain its embedding; (2) If the word is out of vocabulary, we use the averaged embedding of its pieces as its initial node embedding.~The initial word node embeddings are represented as ${\bf X_w} {\rm{=}} {\left[ {{x_{w_1}},{x_{w_2}}, \cdots ,{x_{w_m}}} \right]^{\rm{T}}} \in {\mathbb{R}^{ m \times d}}$.

\noindent \textbf{Category Node Embedding} \  The LIWC\footnote{\url{http://liwc.wpengine.com/}} dictionary divides words into 9 main categories and 64 subcategories.\footnote{Details of the categories are listed in Appendix.}~Empirically, subcategories such as \emph{Pronouns}, \emph{Articles}, and \emph{Prepositions} are not task-related.~Besides, our initial experiments show that excessive introduction of subcategories in the tripartite graph makes the graph sparse and makes the learning difficult, resulting in performance deterioration. For these reasons, we select all 9 main categories and the 6 personal-concern subcategories for our study.~Particularly, the 9 main categories \emph{Function}, \emph{Affect}, \emph{Social}, \emph{Cognitive Processes}, \emph{Perceptual Processes}, \emph{Biological Processes}, \emph{Drives}, \emph{Relativity}, and \emph{Informal Language}, and 6 personal-concern subcategories \emph{Work}, \emph{Leisure}, \emph{Home}, \emph{Money}, \emph{Religion}, and \emph{Death} are 
used as our category nodes.~Then, we replace the ``UNUSED" tokens in BERT's vocabulary by the 15 category names and look up the BERT embedding layer to generate their embeddings ${\bf X_c} {\rm{=}} {\left[ {{x_{c_1}},{x_{c_2}}, \cdots ,{x_{c_n}}} \right]^{\rm{T}}} \in {\mathbb{R}^{ n \times d}}$.

\subsection{Graph Learning}
\begin{figure}[t]
	\centering
	\includegraphics[width=0.45\textwidth]{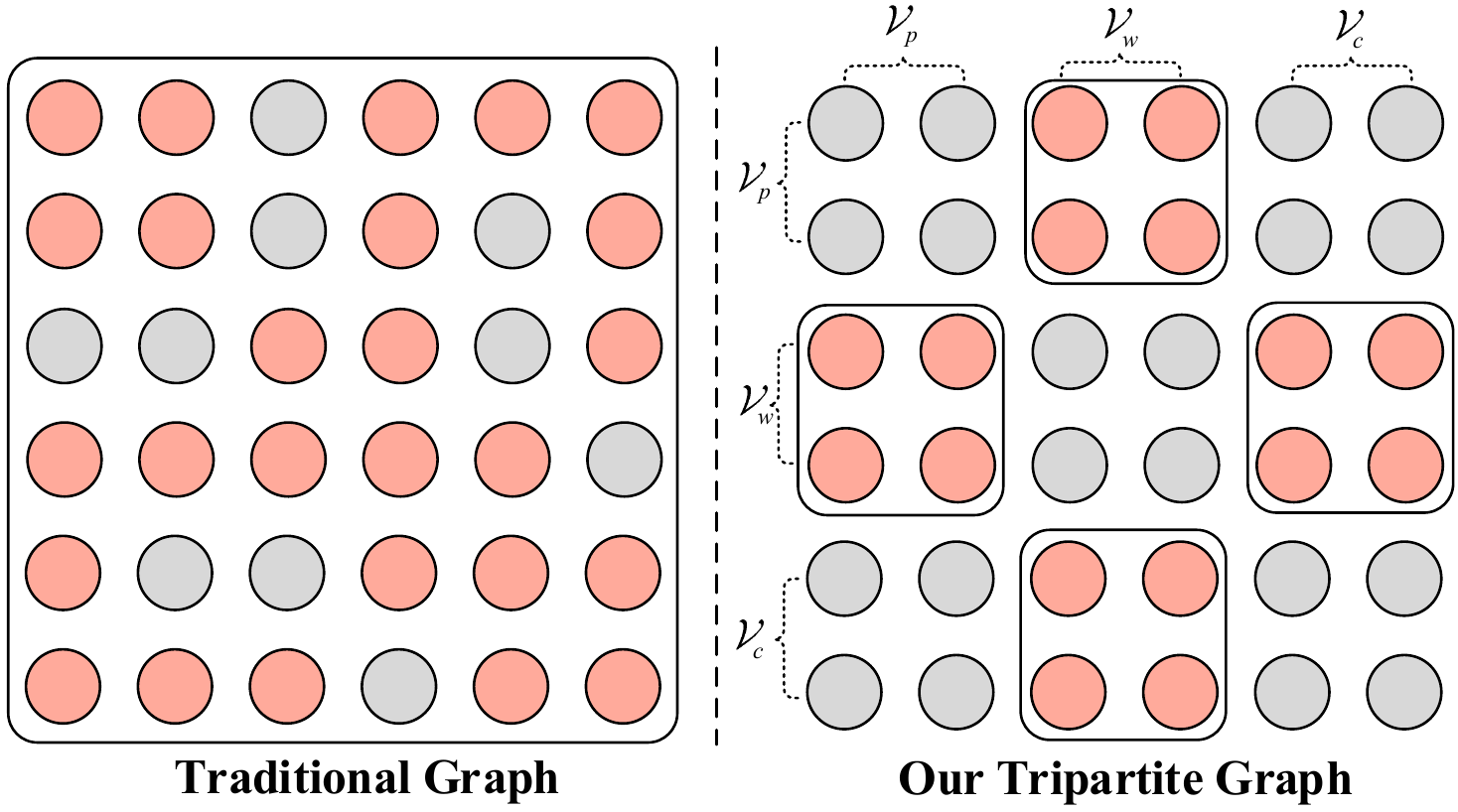}
	\caption{Comparison of adjacent matrices between the traditional graph (left) and our tripartite graph (right). Edges in the traditional graph may occur in any two nodes, while it only occurs between neighboring parties in our tripartite graph.}
	\label{figure3}
\end{figure}

Graph attention network (GAT) \cite{velivckovic2017graph} can be applied over a graph to calculate the attention weight of each edge and update the node representations.~However, unlike the traditional graph in which any two nodes may have edges, the connections in our tripartite graph only occur between neighboring parties (i.e., ${\mathcal V_w} \leftrightarrow {\mathcal V_p}$ and ${\mathcal V_w} \leftrightarrow {\mathcal V_c}$), as shown in Figure \ref{figure3}. Therefore, applying the original GAT over our tripartite graph will lead to unnecessary computational costs. Inspired by \citet{wang2020heterogeneous}, we propose a flow GAT for the tripartite graph. Particularly, considering that the interaction between posts in our tripartite graph can be accounted for by two flows ``$p {\rm{\leftrightarrow}} w {\rm{\leftrightarrow}} p$" and ``$p {\rm{\leftrightarrow}} w {\rm{\leftrightarrow}} c {\rm{\leftrightarrow}} w {\rm{\leftrightarrow}} p$", we design a message passing mechanism that only transmits message by the two flows in the tripartite graph. 

Formally, given a constructed tripartite graph $\mathcal G = \left( {\mathcal V,\mathcal E} \right)$, where $\mathcal V = {\mathcal V_p} \cup {\mathcal V_w} \cup {\mathcal V_c}$, and the initial node embeddings ${\bf X}{\rm{ = }}{\bf X_p} {\rm{ \cup }} {{\bf X_w}} {\rm{ \cup }} {{\bf X_c}}$, we compute ${\bf{H}}_{\bf{p}}^{^{\left( l+1 \right)}}$, ${\bf{H}}_{\bf{w}}^{^{\left( l+1 \right)}}$, and ${\bf{H}}_{\bf{c}}^{^{\left( l+1 \right)}}$ as the hidden states of $\mathcal V_p$, $\mathcal V_w$ and $\mathcal V_c$ at the $(l+1)$-th layer. The flow GAT layer is defined as follows:
\begin{equation}
	{\bf{H}}_{\bf{p}}^{^{\left( {l + 1} \right)}}{\rm{,}}{\bf{H}}_{\bf{w}}^{^{\left( {l + 1} \right)}}{\rm{,}}{\bf{H}}_{\bf{c}}^{^{\left( {l + 1} \right)}}{\rm{ = FGAT}}\left( {{\bf{H}}_{\bf{p}}^{^{\left( l \right)}}{\rm{,}}{\bf{H}}_{\bf{w}}^{^{\left( l \right)}}{\rm{,}}{\bf{H}}_{\bf{c}}^{^{\left( l \right)}}} \right)
\end{equation}
where ${\bf{H}}_{\bf{p}}^{^{\left( 1 \right)}} = {{\bf{X}}_{\bf{p}}}$, ${\bf{H}}_{\bf{w}}^{^{\left( 1 \right)}} = {{\bf{X}}_{\bf{w}}}$, and ${\bf{H}}_{\bf{c}}^{^{\left( 1 \right)}} = {{\bf{X}}_{\bf{c}}}$. The function ${\rm{FGAT}}\left(  \cdot  \right)$  is implemented by the two flows:

\begin{equation} \label{eq4}
\begin{array}{l}
	{\bf{\hat H}}_{{\bf{w}} \leftarrow {\bf{p}}}^{^{\left( l \right)}}{\rm{ = }}{\mathop{\rm MP}\nolimits} \left( {{\bf{H}}_{\bf{w}}^{^{\left( l \right)}},{\bf{H}}_{\bf{p}}^{^{\left( l \right)}}} \right)\\
	{\bf{H}}_{{\bf{ p}} \leftarrow {\bf{w,p}}}^{^{\left( l \right)}} = {\mathop{\rm MP}\nolimits} \left( {{\bf{H}}_{\bf{p}}^{^{\left( l \right)}},{\bf{\hat H}}_{{\bf{w}} \leftarrow {\bf{p}}}^{^{\left( l \right)}}} \right)
\end{array}
\end{equation}
\vspace{-0.5cm}
\begin{equation} \label{eq5}
\begin{array}{l}
	{\bf{H}}_{{\bf{c}} \leftarrow {\bf{w,p}}}^{^{\left( l \right)}} = {\mathop{\rm MP}\nolimits} \left( {{\bf{H}}_{\bf{c}}^{^{\left( l \right)}},{\bf{\hat H}}_{{\bf{w}} \leftarrow {\bf{p}}}^{^{\left( l \right)}}} \right)\\
	{\bf{H}}_{{\bf{ w}} \leftarrow {\bf{c,w,p}}}^{^{\left( l \right)}} = {\mathop{\rm MP}\nolimits} \left( {{\bf{\hat H}}_{{\bf{w}} \leftarrow {\bf{p}}}^{^{\left( l \right)}},{\bf{H}}_{{\bf{c}} \leftarrow {\bf{w,p}}}^{^{\left( l \right)}}} \right)\\
	{\bf{H}}_{{\bf{ p}} \leftarrow {\bf{ w,c,w,p}}}^{^{\left( l \right)}} = {\mathop{\rm MP}\nolimits} \left( {{\bf{H}}_{\bf{p}}^{^{\left( l \right)}},{\bf{H}}_{{\bf{ w}} \leftarrow {\bf{c,w,p}}}^{^{\left( l \right)}}} \right)
\end{array}
\end{equation}
\vspace{-0.5cm}
\begin{equation} \label{eq6}
	\begin{array}{l}
		{\bf{H}}_{\bf{p}}^{^{\left( {l + 1} \right)}} = {\mathop{\rm mean}\nolimits} \left( {{\bf{H}}_{{\bf{ p}} \leftarrow {\bf{w,p}}}^{^{\left( l \right)}},{\bf{H}}_{{\bf{ p}} \leftarrow {\bf{ w,c,w,p}}}^{^{\left( l \right)}}} \right)\\
		{\bf{H}}_{\bf{w}}^{^{\left( {l + 1} \right)}} = {\mathop{\rm mean}\nolimits} \left( {{\bf{\hat H}}_{{\bf{w}} \leftarrow {\bf{p}}}^{^{\left( l \right)}},{\bf{H}}_{{\bf{ w}} \leftarrow {\bf{c,w,p}}}^{^{\left( l \right)}}} \right)\\
		{\bf{H}}_{\bf{c}}^{^{\left( {l + 1} \right)}} = {\bf{H}}_{{\bf{c}} \leftarrow {\bf{w,p}}}^{^{\left( l \right)}}
	\end{array}
\end{equation}
where $ {\rm{\leftarrow}}$ means the message is transmitted from the right nodes to the left nodes, ${\mathop{\rm mean}\nolimits} \left(  \cdot  \right)$ is the mean pooling function, and ${\mathop{\rm MP}\nolimits} \left(  \cdot  \right)$ represents the message passing function.~Eq.~(\ref{eq4}) and Eq.~(\ref{eq5}) illustrate that message is transmitted by the flows ``$p {\rm{\leftrightarrow}} w {\rm{\leftrightarrow}} p$" and $p {\rm{\leftrightarrow}} w {\rm{\leftrightarrow}} c {\rm{\leftrightarrow}} w {\rm{\leftrightarrow}} p$, respectively.

We take ${\mathop{\rm MP}\nolimits} \left( {{\bf{H}}_{\bf{w}}^{^{\left( l \right)}},{\bf{H}}_{\bf{p}}^{^{\left( l \right)}}} \right)$ in Eq.~(\ref{eq4}) as an example to introduce the massage passing function, where ${\bf{H}}_{\bf{w}}^{^{\left( l \right)}} = \left[ {h_{{w_1}}^{\left( l \right)},h_{{w_2}}^{\left( l \right)}, \cdots ,h_{{w_m}}^{\left( l \right)}} \right]$ are used as the attention query and ${\bf{H}}_{\bf{p}}^{^{\left( l \right)}}{\rm{ = }}\left[ {h_{{p_1}}^{\left( l \right)},h_{{p_2}}^{\left( l \right)}, \cdots ,h_{{p_r}}^{\left( l \right)}} \right]$ as the key and value. ${\mathop{\rm MP}\nolimits} \left( {{\bf{H}}_{\bf{w}}^{^{\left( l \right)}},{\bf{H}}_{\bf{p}}^{^{\left( l \right)}}} \right)$ can be decomposed into three steps. First, it calculates the attention weight $\beta^{k}_{ij}$ between node $i$ in $\mathcal V_w$ and its neighbor node $j$ in $\mathcal V_p$ at the $k$-th head:
\begin{equation}
	{z^k_{ij}} = \sigma \left( {{{\bf{W}}^k_{\bf{z}}}\left[ {{{\bf{W}}^k_{\bf{w}}}h_{{w_i}}^{\left( l \right)}||{{\bf{W}}^k_{\bf{p}}}h_{{p_j}}^{\left( l \right)}} \right]} \right)
\end{equation}
\vspace{-0.4cm}
\begin{equation}
	{\beta ^k_{ij}} = \frac{{\exp \left( {{z^k_{ij}}} \right)}}{{\sum\nolimits_{q \in {{\cal N}_i}} {\exp \left( {{z^k_{iq}}} \right)} }}
\end{equation}
where $\sigma$ is the LeakyReLU activation function, ${{\bf{W}}^k_{\bf{z}}}$, ${{\bf{W}}^k_{\bf{w}}}$ and ${{\bf{W}}^k_{\bf{p}}}$ are learnable weights, ${\mathcal N}_i$ means that the neighbor nodes of node $i$ in $\mathcal V_p$, and $||$ is the concatenation operation. Second, the updated hidden state $\tilde h_{{w_i}}^{\left( l \right)}$ is obtained by a weighted combination of its neighbor nodes in $\mathcal V_p$:
\begin{equation}
	\tilde h_{{w_i}}^{\left( {l} \right)} = \mathop {||}\limits_{k = 1}^K \tanh \left( {\sum\limits_{j \in {\mathcal N_i}} {\beta _{ij}^k{\bf{W}}_{\bf{v}}^kh_{{p_j}}^{\left( l \right)}} } \right)
\end{equation}
where $K$ is the number of heads and ${{\bf{W}}^k_{\bf{v}}}$ is a learnable weight matrix. Third, noting that the above steps do not take the information of node $i$ itself into account and to avoid gradient vanishing, we introduce a residual connection to produce the final updated node representation:
\begin{equation}
	\hat h_{{w_i}}^{\left( {l} \right)} = h_{{w_i}}^{\left( l \right)} + \tilde h_{{w_i}}^{\left( {l} \right)}
\end{equation}

\subsection{Merge \& Classification}
After $L$ layers of iteration, we obtain the final node representations ${{{\bf{H}}^{\left( L \right)}}}{\rm{ = }}{{\bf{H}}_{\bf{p}}^{^{\left( L \right)}}} {\rm{ \cup }} {{{\bf{H}}_{\bf{w}}^{^{\left( L \right)}}}} {\rm{ \cup }} {{{\bf{H}}_{\bf{c}}^{^{\left( L \right)}}}}$.~Then, we merge all \emph{post} node representations ${{\bf{H}}_{\bf{p}}^{^{\left( L \right)}}}$ via mean pooling to produce the user representation:
\begin{equation}
	u = {\rm{mean}}\left( {\left[ {h_{{p_1}}^{\left( L \right)},h_{{p_2}}^{\left( L \right)}, \cdots ,h_{{p_r}}^{\left( L \right)}} \right]} \right)
\end{equation}
Finally, we employ $T$ softmax-normalized linear transformations to predict $T$ personality traits. For the $t$-th personality trait, we compute:
\begin{equation}
	{p\left( {y^{ t }} \right)} = {\rm{softmax}}\left( {u{\bf{W}}_{\bf{u}}^t + {\bf{b}}_{\bf{u}}^t} \right)
\end{equation}
where ${{\bf{W}}_{\bf{u}}^t}$ is a trainable weight matrix and ${\bf{b}}_{\bf{u}}^t$ is a bias term. The objective function of our TrigNet model is defined as:
\begin{equation}
	J\left( \theta  \right) = \frac{1}{V}\sum\limits_{v = 1}^V {\sum\limits_{t = 1}^T {\left[ { - y_v^{ t }\log p\left( {y_v^{ t }|\theta } \right)} \right]} } 
\end{equation}
where $V$ is the number of training samples, $T$ is the number of personality traits, $y_v^{t}$ is the true label for the $t$-th trait, and $p( {y_v^{t}|\theta })$ is the predicted probability for this trait under parameters $\theta$.

\section{Experiments}
In this section, we introduce the datasets, baselines, and settings of our experiments. 

\subsection{Datasets}
We choose two public MBTI datasets for evaluations, which have been widely used in recent studies \cite{tadesse2018personality,hernandez2017predicting,majumder2017deep,jiang2019automatic,gjurkovic2020pandora}. The Kaggle dataset\footnote{\url{kaggle.com/datasnaek/mbti-type}} is collected from PersonalityCafe,\footnote{\url{http://personalitycafe.com/forum}} where people share their personality types and discussions about health, behavior, care, etc. There are a total of 8675 users in this dataset and each user has 45-50 posts. Pandora\footnote{\url{https://psy.takelab.fer.hr/datasets/}} is another dataset collected from Reddit,\footnote{\url{https://www.reddit.com/}} where personality labels are extracted from short descriptions of users with MBTI results to introduce themselves. There are dozens to hundreds of posts for each of the 9067 users in this dataset. 

The traits of MBTI include \emph{Introversion} vs. \emph{Extroversion} (\emph{I}/\emph{E}), \emph{Sensing} vs. \emph{iNtuition} (\emph{S}/\emph{N}), \emph{Think} vs. \emph{Feeling} (\emph{T}/\emph{F}), and \emph{Perception} vs. \emph{Judging} (\emph{P}/\emph{J}). Following previous works \cite{majumder2017deep,jiang2019automatic}, we delete words that match any personality label to avoid information leaks. The \emph{Macro-F1} metric is adopted to evaluate the performance in each personality trait since both datasets are highly imbalanced, and \emph{average Macro-F1} is used to measure the overall performance. We shuffle the datasets and split them in a 60-20-20 proportion for training, validation, and testing, respectively. According to our statistics, there are respectively 20.45 and 28.01 LIWC words on average in each post in the two datasets, and very few posts (0.021/0.002 posts per user) are presented as disconnected nodes in the graph. We show the statistics of the two datasets in Table \ref{tab:dataset}. 

\begin{table}[t]
	\centering
	\renewcommand{\arraystretch}{1.1}
	\resizebox{0.485\textwidth}{!}{
		\begin{tabular}{l|cccc}
			\toprule
			\textbf{Dataset} & \textbf{Traits} & \textbf{Train (60\%)} & \textbf{Valid (20\%)} & \textbf{Test (20\%)} \\
			\midrule
			\multirow{4}{*}{Kaggle} 
			& \textit{I}/\textit{E} & 4011 / 1194 & 1326 / 409 & 1339 / 396  \\
			& \textit{S}/\textit{N} & 610 / 4478 & 222 / 1513 & 248 / 1487 \\
			& \textit{T}/\textit{F} & 2410 / 2795 & 791 / 944 & 780 / 955  \\
			& \textit{P}/\textit{J} & 3096 / 2109 & 1063 / 672 & 1082 / 653 \\
			\midrule
			\multirow{4}{*}{Pandora}
			& \textit{I}/\textit{E} & 4278 / 1162 & 1427 / 386 & 1437 / 377  \\
			& \textit{S}/\textit{N} & 727 / 4830 & 208 / 1605 & 210 / 1604 \\
			& \textit{T}/\textit{F} & 3549 / 1891 & 1120 / 693 & 1182 / 632  \\
			& \textit{P}/\textit{J} & 3211 / 2229 & 1043 / 770 & 1056 / 758 \\
			\bottomrule
		\end{tabular}
	}
	\caption{Statistics of the Kaggle and Pandora datasets.}
	\label{tab:dataset}
\end{table}

\begin{table*}[htbp]
	\renewcommand{\arraystretch}{1.0}
	\centering
	\resizebox{1.0\textwidth}{!}{
		\begin{tabular}{l|cccc|c|cccc|c}
			\toprule
			\multirow{2}{*}{\textbf{Methods}}& \multicolumn{5}{c|}{\textbf{Kaggle}}&\multicolumn{5}{c}{\textbf{Pandora}}\\ 
			\cline{2-6}
			\cline{6-11}
			& \textit{\textbf{I}/\textbf{E}} & \textit{\textbf{S}/\textbf{N}} &  \textit{\textbf{T}/\textbf{F}} & \textit{\textbf{P}/\textbf{J}} &  \textbf{Average} & \textit{\textbf{I}/\textbf{E}} & \textit{\textbf{S}/\textbf{N}} &  \textit{\textbf{T}/\textbf{F}} & \textit{\textbf{P}/\textbf{J}} & \textbf{Average} \\
			\midrule
			SVM \cite{cui2017survey}& 53.34 & 47.75 & 76.72 & 63.03 & 60.21 & 44.74 & 46.92 & 64.62 & 56.32 & 53.15 \\
			XGBoost \cite{tadesse2018personality}&  56.67 & 52.85 & 75.42 & 65.94 & 62.72 & 45.99 & 48.93 & 63.51 & 55.55 & 53.50 \\
			BiLSTM \cite{tandera2017personality} & 57.82 & 57.87 & 69.97 & 57.01 & 60.67 & 48.01 & 52.01 & 63.48 & 56.21 & 54.93 \\
			BERT \cite{keh2019myers} & 64.65 & 57.12 & 77.95 & 65.25 & 66.24 & 56.60 & 48.71 & 64.70 & 56.07 & 56.52\\
			AttRCNN \cite{xue2018deep}& 59.74 & 64.08 & 78.77 & 66.44 & 67.25 & 48.55 & \textbf{56.19} & 64.39 & 57.26 & 56.60 \\
			SN+Attn \cite{lynn2020hierarchical}& 65.43 & 62.15 & 78.05 & 63.92 & 67.39  & \textbf{56.98} & 54.78 & 60.95 & 54.81 & 56.88 \\
			\midrule
			TrigNet(our) & \textbf{69.54} & \textbf{67.17} & \textbf{79.06} & \textbf{67.69} & \textbf{70.86} & 56.69 & 55.57 & \textbf{66.38} & \textbf{57.27} & \textbf{58.98}\\
			\bottomrule
	\end{tabular}}
    \caption{Overall results of TrigNet and baselines in \emph{Macro-F1}(\%) score, where the best results are shown in bold.}
    \label{tab:overall}
\end{table*}

\subsection{Baselines}
The following mainstream models are adopted as baselines to evaluate our model:

\noindent \textbf{SVM} \cite{cui2017survey} and \textbf{XGBoost} \cite{tadesse2018personality}: Support vector machine (SVM) or XGBoost is utilized as the classifier with features extracted by TF-IDF and LIWC from all posts.

\noindent \textbf{BiLSTM} \cite{tandera2017personality}:~Bi-directional LSTM \cite{hochreiter1997long} is firstly employed to encode each post, and then the averaged post representation is used for user representation. Glove \cite{pennington2014glove} is employed for the word embeddings.

\noindent \textbf{BERT} \cite{keh2019myers}: The fine-tuned BERT is firstly used to encode each post, and then mean pooling is performed over the post representations to generate the user representation.

\noindent \textbf{AttRCNN}: This model adopts a hierarchical structure, in which a variant of Inception \cite{szegedy2016inception} is utilized to encode each post and a CNN-based aggregator is employed to obtain the user representation. Besides, it considers psycholinguistic knowledge by concatenating the LIWC features with the user representation.

\noindent \textbf{SN+Attn} \cite{lynn2020hierarchical}: As the latest model, SN+Attn employs a hierarchical attention network, in which a GRU \cite{cho2014learning} with word-level attention is used to encode each post and another GRU with post-level attention is used to generate the user representation.

To make a fair comparison between the baselines and our model, we replace the post encoders in AttRCNN and SN+Attn with the pre-trained BERT.

\subsection{Training Details}
We implement our TrigNet in Pytorch\footnote{\url{https://pytorch.org/}} and train it on four NVIDIA RTX 2080Ti GPUs.~Adam \cite{kingma2014adam} is utilized as the optimizer, with the learning rate of BERT set to 2e-5 and of other components set to 1e-3. We set the maximum number of posts, $r$, to 50 and the maximum length of each post, $s$, to 70, considering the limit of available computational resources. After tuning on the validation dataset, we set the dropout rate to 0.2 and the mini-batch size to 32. The maximum number of nodes, $r+m+n$, is set to 500 for Kaggle and 970 for Pandora, which cover 98.95\% and 97.07\% of the samples, respectively. Moreover, the two hyperparameters, the numbers of flow GAT layers $L$ and heads $K$, are searched in $\{1,2,3\}$ and $\{1,2,4,6,8,12,16,24\}$, respectively, and the best choices are $L=1$ and $K=12$. The reasons for $L=1$ are likely twofold. First, our flow GAT can already realize the interactions between nodes when $L=1$, whereas the vanilla GAT needs to stack 4 layers. Second, after trying $L=2$ and $L=3$, we find that they lead to slight performance drops compared to that of $L=1$.

\section{Results and Analyses}
In this section, we report the overall results and provide thorough analyses and discussions.

\subsection{Overall Results}
The overall results are presented in Table \ref{tab:overall}, from which our observations are described as follows. First, the proposed TrigNet consistently surpasses the other competitors in F1 scores, demonstrating the superiority of our model on text-based personality detection with state-of-the-art performance. Specifically, compared with the existing state of the art, SN+Attn, TrigNet achieves 3.47 and 2.10 boosts in average F1 on the Kaggle and Pandora datasets, respectively.~Second, compared with BERT, a basic module utilized in TrigNet, TrigNet yields 4.62 and 2.46 improvements in average F1 on the two datasets, verifying that the tripartite graph network can effectively capture the psychological relations between posts. Third, compared with AttRCNN, another method of leveraging psycholinguistic knowledge, TrigNet outperforms it with 3.61 and 2.38 increments in average F1 on the two datasets, demonstrating that our solution that injects psycholinguistic knowledge via the tripartite graph is more effective. Besides, the shallow models SVM and XGBoost achieve comparable performance to the non-pre-trained model BiLSTM, further showing that the words people used are important for personality detection.

\begin{table}[t]
\small
	\renewcommand{\arraystretch}{1.1}
	\centering
	\resizebox{0.485\textwidth}{!}
	{
		\begin{tabular}{lcc}
			\toprule
			\textbf{Model}& \textbf{Ave. F1(\%)}&$\bm{\Delta}$\textbf{(\%)}\\ 
			\midrule
			TrigNet & 70.86 & - \\
			\midrule
			w/o ``$p {\rm{\leftrightarrow}} w {\rm{\leftrightarrow}} p$" & 70.13 & 0.73$\downarrow$  \\
			w/o``$p {\rm{\leftrightarrow}} w {\rm{\leftrightarrow}} c {\rm{\leftrightarrow}} w {\rm{\leftrightarrow}} p$" & 69.56 & 1.3$\downarrow$   \\
			w/o Layer attention & 69.88 & 0.98$\downarrow$   \\
			\midrule
			w/o Function  & 70.44 & 0.42$\downarrow$  \\
			w/o Perceptual processes & 70.28 & 0.58$\downarrow$  \\
			w/o Work & 70.28 &  0.58$\downarrow$ \\
			w/o Home & 70.08 &  0.78$\downarrow$ \\
			w/o Drives & 70.03 &  0.83$\downarrow$ \\
			w/o Relativity & 69.91 &  0.95$\downarrow$ \\
			w/o Cognitive processes & 69.69 &  1.17$\downarrow$ \\
			w/o Biological processes & 69.68 &  1.18$\downarrow$ \\
			w/o Leisure & 69.67 &  1.19$\downarrow$ \\
			w/o Religion & 69.58 &  1.28$\downarrow$ \\
			w/o Money & 69.56 &  1.30$\downarrow$ \\
			w/o Informal language & 69.51 & 1.35$\downarrow$  \\
			w/o Social & 69.32 &  1.54$\downarrow$ \\
			w/o Death & 69.30 &  1.56$\downarrow$ \\
			w/o Affect & 68.60 &  2.26$\downarrow$ \\
			\bottomrule
	\end{tabular}}
    \caption{Results of ablation study in average Macro-F1 on the Kaggle dataset, where ``w/o'' means removal of a component from the original TrigNet, and ``$\Delta$'' indicates the corresponding performance change.}
    \label{tab:ablation}
    		\vspace{-0.3cm}
\end{table}

\begin{figure*}[hbp]
	\centering
	\includegraphics[width=0.98\textwidth]{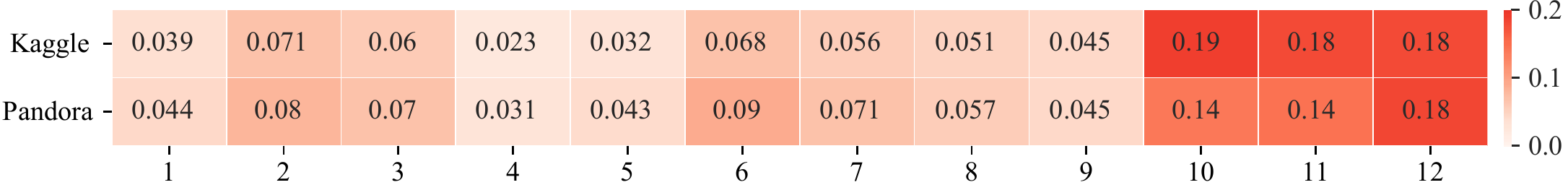}
	\caption{Visualization of layer attention weights. The last three layers supply with more information for this task.}
	\label{fig:attention_weight}
\end{figure*}

\subsection{Ablation Study}
We conduct an ablation study of our TrigNet model on the Kaggle dataset by removing each component to investigate their contributions.~Table \ref{tab:ablation} shows the results which are categorized into two groups. 

In the first group, we investigate the contributions of the network components. We can see that removing the flow ``$p {\rm{\leftrightarrow}} w {\rm{\leftrightarrow}} c {\rm{\leftrightarrow}} w {\rm{\leftrightarrow}} p$" defined in Eq. (\ref{eq5}) results in higher performance declines than removing the flow ``$p {\rm{\leftrightarrow}} w {\rm{\leftrightarrow}} p$" defined in Eq. (\ref{eq4}), implying that the \emph{category} nodes are helpful to capture personality cues from the texts. Besides, removing the layer attention mechanism also leads to considerable performance degradation.

In the second group, we investigate the contribution of each category node. The results, sorted by scores of decrease from small to large, demonstrate that the introduction of every category node is beneficial to TrigNet. Among these category nodes, the \emph{Affect} is shown to be the most crucial one to our model, as the average Macro-F1 score drops most significantly after it is removed. This implies that the \emph{Affect} category reflects one's personality obviously.~Similar conclusions are reported by \citet{depue1999neurobiology} and \citet{zhang2019persemon}.
In addition, the \emph{Function} node is the least impactful category node. The reason could be that functional words reflect pure linguistic knowledge and are weakly connected to personality.

\vspace{-0.1cm}
\subsection{Analysis of the Computational Cost}
In this work we propose a flow GAT to reduce the computational cost of vanilla GAT. To show its effect, we compare it with vanilla GAT (as illustrated in the left part of Figure \ref{figure3}). The results are reported in Table \ref{tab:cost}, from which we can observe that flow GAT successfully reduces the computational cost in FLOPS and Memory by 38\% and 32\%, respectively, without extra parameters introduced. Besides, flow GAT is superior to vanilla GAT when the number of layers is 1. The cause is that the former can already capture adequate interactions between nodes with one layer, while the latter has to stack four layers to achieve this.

We also compare our TrigNet with the vanilla BERT in terms of the computational cost. The result show that the flow GAT takes about 1.14\% more FLOPS than the vanilla BERT(297.3G).

\begin{table}[t]
	\renewcommand{\arraystretch}{1.0}
	\centering
	\resizebox{0.48\textwidth}{!}
	{
		\begin{tabular}{lccccc}
			\toprule
			\textbf{GAT}& \textbf{Params}& \textbf{FLOPS} & \textbf{Memory}  & \textbf{Ave.F1} \\ 
			\midrule
			Original & 1.8M & 5.5G & 7.8GB & 69.69 \\
			Flow(our) & 1.8M & 3.4G & 5.3GB & 70.86 \\
			\bottomrule
	\end{tabular}}
	\caption{Analysis of the computational cost for original GAT and flow GAT on the Kaggle dataset.~The metrics include the number of parameters (Params) and floating-point operations per second (FLOPS) of GAT as well as memory size (Memory) and the average Macro-F1 (Ave.F1) of whole model on the Kaggle dataset.}
	\label{tab:cost}
	\vspace{-0.1cm}
\end{table}

\subsection{Layer Attention Analysis}
\label{sec:layer_attention}
This study adopts layer attention \cite{peters2018deep} as shown in Eq. (\ref{eq2}) to produce initial embeddings for post nodes. To show which layers are more useful, we conduct a simple experiment on the two datasets by using all the 12 layer representations of BERT and visualize the attention weight of each layer. As plotted in Figure \ref{fig:attention_weight}, we find that the attention weights from layers 10 to 12 are significantly greater than that of the rest layers on both datasets, which explains why the last three layers are chosen for layer attention in our model.

\section{Conclusion} 
In this work, we proposed a novel psycholinguistic knowledge-based tripartite graph network, TrigNet, for personality detection. TrigNet aims to introduce structural psycholinguistic knowledge from LIWC via constructing a tripartite graph, in which interactions between posts are captured through psychologically relevant words and categories rather than simple document-word or sentence-word relations. Besides, a novel flow GAT that only transmits messages between neighboring parties was developed to reduce the computational cost. Extensive experiments and analyses on two datasets demonstrate the effectiveness and efficiency of TrigNet. This work is the first effort to leverage a tripartite graph to explicitly incorporate psycholinguistic knowledge for personality detection, providing a new perspective for exploiting domain knowledge.

\section*{Acknowledgments}
The paper was fully supported by the Program for Guangdong Introducing Innovative and Entrepreneurial Teams (No.2017ZT07X355).

\section*{Ethical Statement} 
This study aims to develop a technical method to incorporate psycholinguistic knowledge into neural models, rather than creating a privacy-invading tool. We worked within the purview of acceptable privacy practices and strictly followed the data usage policy. The datasets used in this study are all from public sources with all user information anonymized. The assessment results of the proposed model are sensitive and should be shared selectively and subject to the approval of the institutional review board (IRB). Any research or application based on this study is only allowed for research purposes, and any attempt to use the proposed model to infer sensitive user characteristics from publicly accessible data is strictly prohibited.

\bibliographystyle{acl_natbib}
\bibliography{anthology,acl2021}

\appendix

\begin{figure*}[b]
	\centering
	\includegraphics[width=1.0\textwidth]{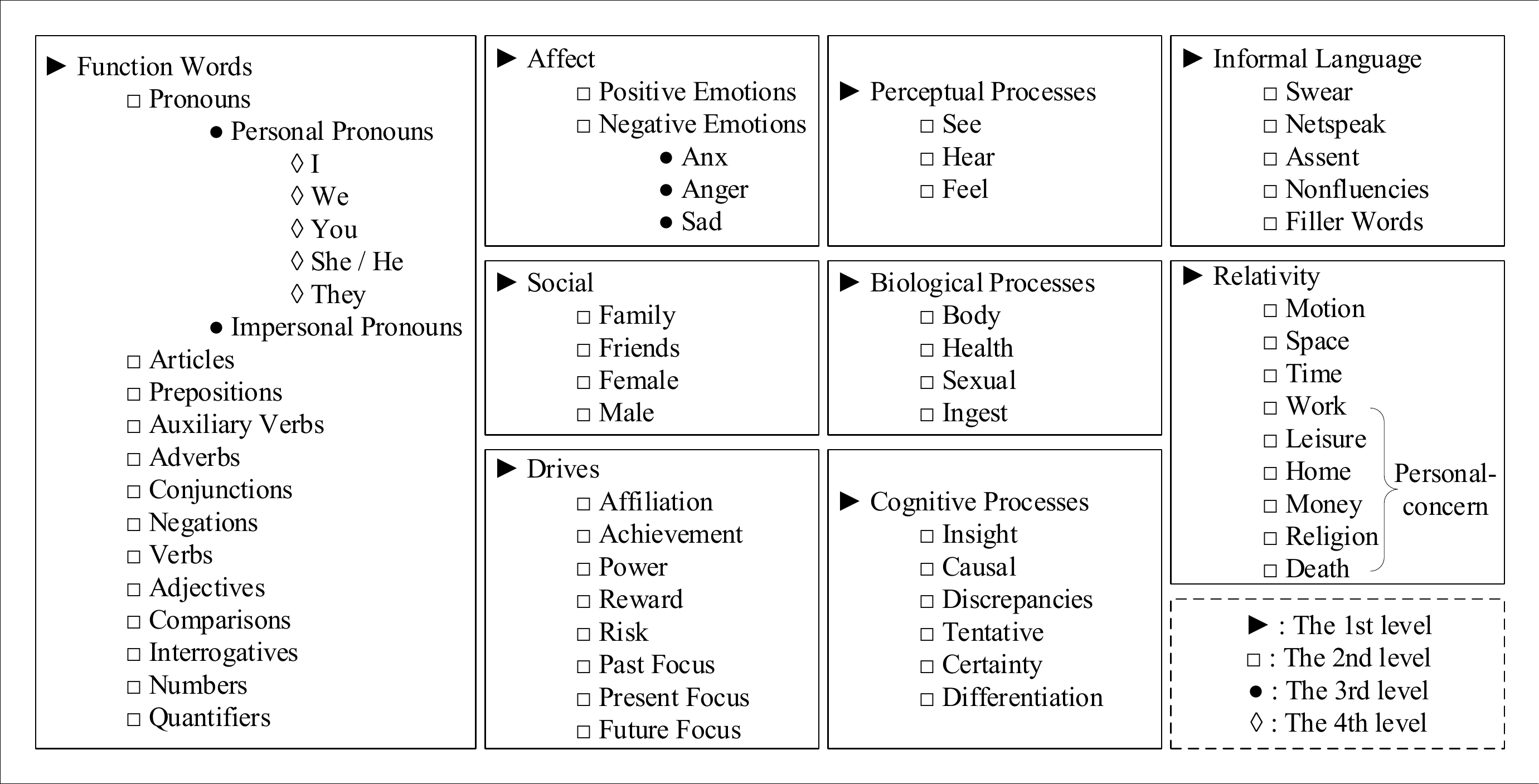}
	\caption{Detailed division of categories in the LIWC-2015 dictionary.}
	\label{fig:categories}
\end{figure*}

\section{Categories of LIWC }
As shown in Figure \ref{fig:categories}, a total of 73 categories and subcategories are defined in the LIWC-2015 dictionary. There are 9 main categories: \emph{Function}, \emph{Affect}, \emph{Social}, \emph{Cognitive Processes}, \emph{Perceptual Processes}, \emph{Biological Processes}, \emph{Drives}, \emph{Relativity}, and \emph{Informal Language}, in which 20 standard linguistic subcategories are included in the \emph{Function} category and 44 psychological-relevant subcategories are defined in the rest 8 categories.

\end{document}